\pdfoutput=1

\documentclass[11pt]{article}
\usepackage{graphicx}
\usepackage{subcaption}
\usepackage{booktabs}
\usepackage{listings}
\usepackage{setspace}
\usepackage{framed}
\usepackage[preprint]{acl}

\usepackage{times}
\usepackage{latexsym}
\usepackage{url}
\usepackage[utf8]{inputenc}
\usepackage[T1]{fontenc}

\usepackage[utf8]{inputenc}

\usepackage{microtype}

\usepackage{inconsolata}

\usepackage{graphicx}
\usepackage{booktabs}
\usepackage{textcomp}
\title{%
\includegraphics[width=0.05\textwidth]{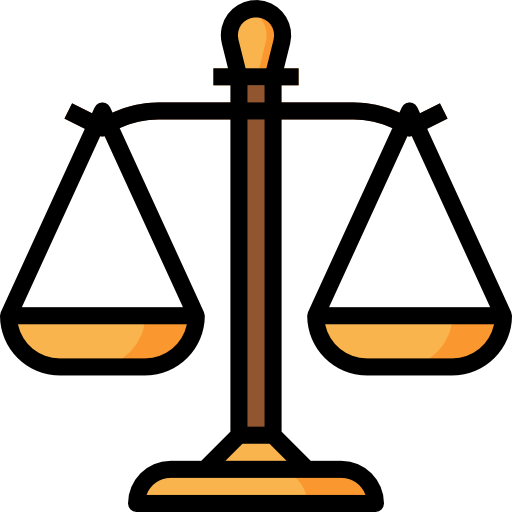} 
\includegraphics[width=0.05\textwidth]{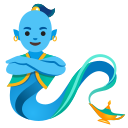}  
  LexGenie: Automated Generation of Structured Reports for European Court of Human Rights Case Law%
}

\author{ Santosh T.Y.S.S$^{1}$, Mahmoud Aly$^1$, Oana Ichim$^2$, Matthias Grabmair$^1$
\\ $^1$School of Computation, Information, and Technology; \\ Technical University of Munich, Germany \\
$^2$Graduate Institute of International and Development Studies, Geneva, Switzerland\\ 
 \texttt{\{santosh.tokala, mahmoud.aly, matthias.grabmair\}@tum.de} \\
  \texttt{oana.ichim@graduateinstitute.ch} }

\begin{document}
\maketitle
\begin{abstract}
Analyzing large volumes of case law to uncover evolving legal principles, across multiple cases, on a given topic is a demanding task for legal professionals. Structured topical reports provide an effective solution by summarizing key issues, principles, and judgments, enabling comprehensive legal analysis on a particular topic. While prior works have advanced query-based individual case summarization, none have extended to automatically generating multi-case structured reports. To address this, we introduce LexGenie, an automated LLM-based pipeline designed to create structured reports using the entire body of case law on user-specified topics within the European Court of Human Rights jurisdiction. LexGenie retrieves, clusters, and organizes relevant passages by topic to generate a structured outline and cohesive content for each section. Expert evaluation confirms LexGenie's utility in producing structured reports that enhance efficient, scalable legal analysis.
\end{abstract}

\section{Introduction}
Court judgments, beyond resolving individual cases, play a critical role in developing, clarifying, and safeguarding legal principles, ensuring the consistent application of law within a given jurisdiction \cite{farzindar2004atefeh,saravanan2006improving,santosh2025coperlex}. Consequently, legal professionals face the challenging task of analyzing and synthesizing large volumes of complex case law to extract relevant legal precedents, understand the application of laws, and inform their legal strategies \cite{bhattacharya2019comparative,tyss2024beyond,santosh2025relexed}. In response to this growing demand, recent efforts have focused on automatic summarization of individual cases, which condense the content of a single case, making it easier for legal professionals to quickly grasp key points \cite{zhong2019automatic,shukla2022legal,deroy2023ready,santosh2024lexsumm}. In practice, a single case may include multiple documents of varied types, such as complaints, opinions, motions, briefs, settlements, affidavits, and discovery materials—often totaling hundreds of pages per case, leading to exploration of multi-document legal summarization systems that process and distill information across multiple legal texts \cite{shen2022multi}. Furthermore, a one-size-fits-all approach that produces a single, generic summary may not sufficiently address the diverse and specific needs of legal professionals and this limitation has spurred the development of aspect or query-focused case summarization systems, which provide tailored summaries based on users' specific information needs, allowing for a more customized and relevant output \cite{tyss2024lexabsumm}.

\begin{figure}[!t]
    \centering
  \includegraphics[width=0.9\linewidth]{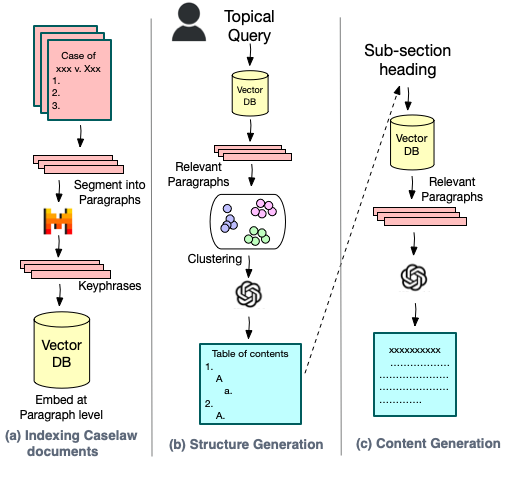}
    \caption{Overview of our approach, LexGenie.}
    \label{fig:lexgenie}
\end{figure}

While these solutions represent significant progress, they remain limited to single-case contexts, often missing the broader perspective necessary to track the evolution of legal principles across multiple cases. For more strategic analysis, legal professionals require cross-case insights that reveal how precedents and interpretations develop over time. To meet this need, structured reports are typically prepared, focusing on specific Articles or Transversal Themes. These reports summarize key principles, issues, and judgments drawn from multiple cases by identifying relevant cases, recognizing common legal patterns, and and organizing the information into a multidimensional framework—akin to a detailed table of contents, with insights structured under each dimension.  Yet, manually generating these structured, multi-case reports is labor-intensive and time-consuming. As case law grows in volume and legal issues increase in complexity, the demand for automatically creating these reports has become pressing. Our work addresses this challenge by moving beyond single-case summarization toward an extreme summarization approach that synthesizes patterns and principles across entire body of case law. We explore the utility of current technologies, such as large language models (LLMs), to assist in generating structured reports that support a comprehensive, cross-case understanding of key legal issues.

We develop LexGenie, a fully automated pipeline leveraging LLMs to generate structured reports based on the topical queries issued by the user, focusing on European Court of Human Rights (ECHR) Jurisdiction, which adjudicates complaints by individuals against states about alleged violations of their rights as enshrined in the European Convention of Human Rights. LexGenie employs a two-stage pipeline: in the first stage, it retrieves relevant passages according to the user’s query, optimizing for recall and performs clustering to create a topic-based outline for the report. In the second stage, LexGenie generates  content for each section by sourcing precise paragraphs for that sub-topic. We validate LexGenie's effectiveness through a small-scale evaluation conducted by an ECHR legal expert, demonstrating its ability to produce accurate and relevant report structure and content. Additionally, we examine whether LLMs can assist in assessing output quality, finding a positive correlation with expert annotations.

\section{LexGenie}
We present the methodology behind LexGenie, for generating structured reports from ECHR case law judgments based on user-issued topical queries. LexGenie structurally organizes each report into a coherent dimensions related to the topic, enabling users to navigate and understand a thematic legal area, supported by references to relevant case law judgments. We then describe our user interface, designed for accessibility and ease of use.

\subsection{Approach}
LexGenie’s workflow comprises three main steps: (i) indexing case law documents at the paragraph level, offline, into a vector datastore for efficient query-based retrieval, (ii) structure generation module, which retrieves relevant paragraphs based on the query, organizes them into hierarchical thematic clusters to finally generate a coherent outline with headings and sub-headings and (iii) content generation, where relevant content is sourced and expanded upon for each subsection of the outline.

\subsubsection{Indexing Case law documents} 
We gather the complete ECHR case law collection from the latest version of \citet{santosh2024ecthr,santosh2025lecopcr}, sourced from HUDOC, the public ECHR database. Each judgment is organized by paragraph numbers, which serve as the primary unit for cross-referencing within the ECHR writing style.

Rather than indexing the raw paragraph text as embeddings, we use a keyphrase-based approach to represent each paragraph’s main themes. This focus on keyphrases enhances the embeddings by centering them around key legal concepts, while minimizing the inclusion of case-specific details that would otherwise arise with full-text embeddings, thus facilitating accurate, thematic matches with user queries. To obtain these keyphrases, we prompt the Mistral-7B-Instruct model \cite{jiang2023mistral}\footnote{\url{https://huggingface.co/mistralai/Mistral-7B-Instruct-v0.3}} using each paragraph's text. Appendix \ref{keyphrase} provides the prompt and an example of paragraph-level keyphrase generation. To improve efficiency, we use batch prompting \cite{cheng2023batch}, running inference in groups of paragraphs sourced from the same judgment document rather than one at a time. This approach reduces token and processing time costs while contextualizing each paragraph within the broader scope of the case. Once generated, these paragraph keywords are concatenated and embedded using OpenAI’s text-embedding-3-small model. We store the resulting dense vector embeddings in a FAISS database, integrated via the LangChain framework\footnote{\url{https://www.langchain.com/langchain}}, which enables efficient, semantically-similar retrieval.

\subsubsection{Structure Generation}
This module analyzes the entire body of case law to extract relevant concepts related to user queries and organizes them into a coherent table of contents. By structuring sub-topics effectively, it enhances the user's understanding of key legal themes and facilitates navigation through complex subjects. The process involves four main steps: retrieving relevant paragraphs, hierarchically clustering them based on shared themes, generating topical headings for each cluster, and organizing these headings into a cohesive narrative flow.

First, we retrieve relevant paragraphs based on Maximal Marginal Relevance (MMR) \cite{carbonell1998use} using the LangChain framework. MMR balances  relevance (semantic similarity with query) and diversity (semantic similarity between retrieved items), ensuring that the selected paragraphs encompass a broad spectrum of themes related to the query topic. Next, we apply BERTopic \cite{grootendorst2022bertopic} to cluster the retrieved paragraphs, which helps in identifying and organizing common themes. Utilizing the text-embedding-3-small model in conjunction with HDBSCAN, we generate hierarchical topical clusters \cite{mcinnes2017hdbscan}. To create topic headings for each cluster, we prompt the GPT-4o-mini model with five representative paragraphs from each cluster, as detailed in Appendix \ref{strcuture}. Once the topic names are generated for each cluster individually, we finally prompt GPT-4o-mini to refine all the headings and subheadings into a cohesive, ordered structure. This can involve reordering, merging and organizing topics to ensure logical flow across all (sub-)clusters, resulting in a well-structured report outline. Detailed prompt is provided in App. \ref{strcuture}.

\subsubsection{Content Generation}
In this phase, we generate content for each sub-section (leaf node) in the established table of contents. First, we construct a query by concatenating the sub-heading with the headings along its hierarchical path from the root node and is used to retrieve the top relevant paragraphs from the datastore. This augmented query, providing contextual relevance enables the retrieval of more precise paragraphs targeted towards the specific sub-section.

Next, we generate the content for each sub-section using the retrieved paragraphs following an iterative incremental updating approach using the GPT-4o-mini model \cite{chang2023booookscore}, to handle cases where the length of relevant paragraphs exceeds the model's prompt length. In the initial iteration, the model is prompted with 25 relevant paragraphs to generate content for the specified sub-section, while also including references to the corresponding paragraphs. In subsequent iterations, the model receives the content generated up to that point along with the next set of 25 relevant paragraphs, prompting it to modify the previously generated content by integrating any additional insights from the latest paragraphs. Appendix \ref{content_gen} provides the detailed prompts.

\begin{figure*}[h!]
    \centering
    \fbox{\includegraphics[width=0.7\linewidth]{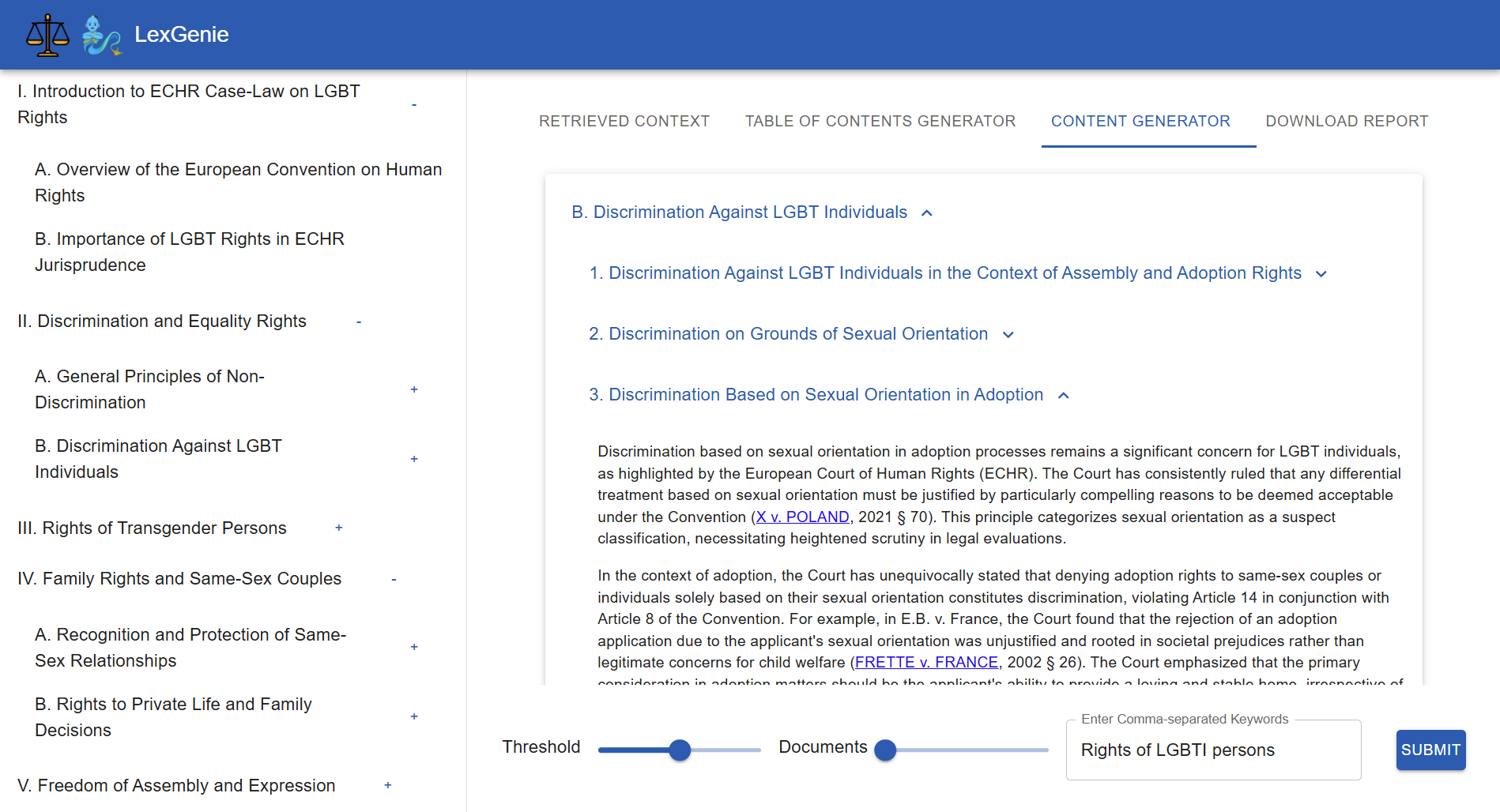}}

    \caption{LexGenie interface. Given a legal topic as query, it automatically retrieves relevant documents and generates a table of content structure for the report. Finally, content for each sub-section in report is populated and the whole report is available for download.}
    \label{fig:three_images}
\end{figure*}

\subsection{User Interface}
LexGenie is accessible as a web app, which can be run locally and is available at \url{https://tinyurl.com/2a9jhrpu}. A video demonstration of LexGenie is available at \url{https://tinyurl.com/585h53cj}. An user inputs a search query to initiate the retrieval process.  Adjustable parameters, such as the number of judgments retrieved and a similarity threshold, allow users to control the scope of retrieved content. In the initial retrieval step, relevant paragraphs are displayed as judgments, with paragraph numbers linked to the original HUDOC case law documents for easy reference. Users can further refine these results by adjusting the ranked list of retrieved items, removing, or adding new passages including additional passages from other cases in the datastore can be incorporated through a fuzzy search-based dropdown menu. Based on these refined paragraphs, a structured table of contents is generated through clustering and organization using LLM calls.

Users can review and edit the generated table of contents before proceeding to content generation. The table of contents is displayed in a side navigation panel, allowing users to navigate through the hierarchy of headings and subheadings. Then users can generate content for individual (sub-)sections or for the entire table of contents by clicking the appropriate buttons. The generated content is post-processed to include citations linking each segment of the report to the respective ECHR documents, based on the references provided by the LLM model. The final report is available for download in PDF format. LexGenie’s UI is designed to reflect the underlying pipeline, making each step in the report creation process transparent and customizable. This design enables feedback collection from users, allowing us to assess the interface's effectiveness at each stage of the process.

\section{Evaluation}
\subsection{Report Structure}
We assess the quality of the generated report structure across the following dimensions: (a) Topical Relevance: Emphasizes how closely the generated headings and subheadings align with the user’s query. (b) Subtopic Consistency: Focuses on the alignment of subtopics under each parent heading, ensuring intra-cluster consistency. (c) Cluster Distinction: Highlights the uniqueness of each topic cluster, ensuring clear differentiation and minimal inter-cluster redundancy. (d) Narrative Flow: Evaluates the logical progression of the structure, ensuring it guides the reader smoothly through the topics. (e) Comprehensiveness of Topics: Measures the extent to which the headings and subheadings encompass all critical aspects of the query, avoiding any significant gaps.

\begin{table*}[h]
\setlength{\tabcolsep}{4pt} 
\centering
\scalebox{0.86}{
\begin{tabular}{l|cc|cc|cc|cc|cc}
\hline
\textbf{Model} & \multicolumn{2}{c|}{\textbf{Topical Rel.}} & \multicolumn{2}{c|}{\textbf{Subtopic Con.}} & \multicolumn{2}{c|}{\textbf{Cluster Dist.}} & \multicolumn{2}{c|}{\textbf{Narr. Flow}} & \multicolumn{2}{c}{\textbf{Comprehen.}} \\ 
               & \textbf{Human} & \textbf{Auto} & \textbf{Human} & \textbf{Auto} & \textbf{Human} & \textbf{Auto} & \textbf{Human} & \textbf{Auto} & \textbf{Human} & \textbf{Auto} \\ \hline
LexGenie       & \textbf{3.95}           & \textbf{4.48}          & \textbf{3.75}           & \textbf{4.49}          & \textbf{3.70}           & \textbf{3.91}          & 3.75           & \textbf{4.27}          & \textbf{3.25}           & \textbf{3.27}          \\ 
Paragraph-based & 3.85          & 4.38          & 3.50           & 4.33          & 3.45           & 3.40          & 3.60           & 4.18          & 3.10           & 3.16          \\ 
w/o MMR        & 3.90           & 4.31          & 3.70           & 4.09          & 3.55           & 3.53          & \textbf{3.85}           & 4.22          & 3.0           & 2.84          \\ 
w/o Reorganization & 3.85       & 4.45          & 3.55           & 3.97          & 3.45           & 2.69          & 3.55           & 3.37          & 3.20           & 3.06          \\ \hline
\end{tabular}}
\caption{Human and Automatic Evaluation Results for Report Structure Quality.}
\label{tab:con_str}
\end{table*}

We investigate the effect of each design choice in LexGenie: (i) Keyphrase vs. Paragraph-Based Indexing: While LexGenie employs a keyphrase-based approach to index each paragraph for retrieval and clustering, we modify it to use the raw paragraph text for indexing and further clustering. (ii) Retrieval Strategy: LexGenie uses the Maximal Marginal Relevance (MMR) criterion to balance relevance and diversity. We replace it with a traditional relevance-based criterion. (iii) Impact of Reorganization: LexGenie employs an LLM call to order the generated headings and subheadings into a cohesive structure. We remove this call and concatenate the cluster-based individually generated headings to form the final structure.

\noindent \textbf{Human Evaluation}
We randomly select 20 queries covering broad topics such as Articles and Themes from existing ECHR case law guides\footnote{Case-law guides are structured reports maintained by courts registry, accessible on the ECHR Knowledge Sharing Platform at \url{https://ks.echr.coe.int/web/echr-ks/all-case-law-guides}.}. We then ask a legal expert, the third author of this paper, to manually evaluate the quality of the generated report structures using a 1-5 scale, where a higher score indicates better quality. The evaluation is based on each of the five dimensions outlined above for LexGenie and the three ablation systems.

From Table \ref{tab:con_str},  we observe that the report structure generated by LexGenie is highly rated by our legal expert, reflecting topically relevant headings, well-grouped sub-topics with clear delineation, logically organized narrative flow, and comprehensive coverage of relevant aspects. The keyphrase-based approach significantly outperforms the paragraph-based approach across all metrics, particularly in sub-topic consistency and cluster distinction. This suggests that keyphrase generation effectively steers the model to focus on core legal concepts, while paragraph embeddings tend to capture additional case-specific details, which may dilute relevance in retrieval and clustering.

\begin{table*}[t]
\setlength{\tabcolsep}{3pt} 
\centering
\scalebox{1.0}{
\begin{tabular}{l|cc|cc|cc|cc}
\hline
& \multicolumn{2}{c|}{\textbf{Topical Relevance}}              & \multicolumn{2}{c|}{\textbf{Content Org.}}           & \multicolumn{2}{c|}{\textbf{Citation Faith.}}          & \multicolumn{2}{c}{\textbf{Comprehen.}}              \\
& \textbf{Human}                   & \multicolumn{1}{c|}{\textbf{Auto}} & \textbf{Human}                   & \multicolumn{1}{c|}{\textbf{Auto}} & \textbf{Human}                   & \multicolumn{1}{c|}{\textbf{Auto}} & \textbf{Human}                   & \multicolumn{1}{c}{\textbf{Auto}} \\ \hline
\multicolumn{1}{c|}{Single} & \multicolumn{1}{c}{4.6} & 4.87                     & \multicolumn{1}{c}{\textbf{4.5}} & 4.47                     & \multicolumn{1}{c}{3.9} & 4.29                     & \multicolumn{1}{c}{4.0}   & 4.23                     \\
\multicolumn{1}{c|}{Incremental}          & \multicolumn{1}{c}{\textbf{4.9}} & \textbf{4.94}                     & \multicolumn{1}{c}{\textbf{4.5}} & \textbf{4.59}                     & \multicolumn{1}{c}{\textbf{4.5}} & \textbf{4.36}                     & \multicolumn{1}{c}{\textbf{4.5}} & \textbf{4.55}                    \\ \hline
\end{tabular}}
\caption{Human and Automatic Evaluation Results for Content Quality.}
\label{tab:con_gen}
\end{table*}

When diversity criterion in retrieval (w/o MMR) is removed, we observe the appearance of similar sub-topics among the top retrieved results, leading to gaps in topic coverage, as reflected in lower comprehensiveness scores. The reduced cluster distinction can be attributed to the lack of sub-topic diversity, which complicates clear separation between clusters. Although narrative flow improves slightly due to less diverse sub-topics, this comes at the cost of thematic variety. Lastly, omitting the LLM reorganization step results in declines across narrative flow, sub-topic consistency and cluster distinction. Without reorganization, the structure lacks coherence and topics are less clearly differentiated, ultimately hindering thematic clarity.

\noindent \textbf{Automatic Evaluation} We evaluate the capabilities of LLMs to conduct automated assessments across the five dimensions using the same set of 20 queries selected for human evaluation.  We employ the G-Eval \cite{liu2023g} framework, which prompts LLMs with chain-of-thought and a form-filling paradigm, to assess the quality of generated outputs. For all metrics, we provide the detailed instruction, generated report structure along with the query to provide an assessment in scale of 1-5. For comprehensiveness of topics evaluation, which requires additional external knowledge to understand the topical coverage, we also provide the model with table of content structure from original case law guides as reference context. This allows the model to compare the generated content structure against this reference to identify the missing aspects, to assess comprehensiveness.

From Table \ref{tab:con_str}, we observe that LexGenie achieves high scores across automated metrics, aligning closely with human expert evaluations. Notably, the automated metrics reveal lower comprehensiveness scores for the approach without MMR, attributed to reduced sub-topic diversity in the retrieval process—an observation mirrored in the expert assessments. Likewise, the absence of reorganization adversely impacts narrative flow and both intra- and inter-cluster consistency. Additionally, the paragraph-based approach underperforms relative to the keyphrase-based approach, both in retrieval and clustering, suggesting that keyphrase-based representations better capture core topics enhancing intra- and inter- cluster consistency.

\subsection{Content Generation}
We assess the quality of the generated content under each (sub-)heading across the following dimensions: (a) Topical Relevance: measures how well the generated content aligns with the (sub-)section heading. 
(b) Content Organization: evaluates the logical flow and coherence of the content throughout.
(c) Citation Faithfulness:  assesses the extent to which the generated content is supported by appropriate and reliable citations. (d) Comprehensiveness: examines whether all relevant aspects of the section topic are comprehensively addressed, ensuring no critical information is overlooked. While an incremental prompting is used in LexGenie, we compare it with a single prompting approach where content is generated using all retrieved passages provided to the model simultaneously.

\noindent \textbf{Human Evaluation}
We randomly select 10 (sub-)headings from existing ECHR case law guides, which serve as leaf nodes and generate corresponding content for each.  A legal expert manually evaluates the quality of the generated content for each heading on a 1-5 scale across the four dimensions, with higher scores indicating better quality. As shown in Table \ref{tab:con_gen}, both the incremental and single prompting approaches maintain a coherent narrative structure. However, the incremental prompting generates content that is more firmly grounded in the provided heading and retrieved paragraphs, with appropriate citations, in contrast to the single prompting approach. The lower performance of the single setup can be attributed to the overwhelming amount of content presented to the LLMs, which complicates the distillation of important information across multiple paragraphs. This suggests that the model is better able to focus on relevant aspects when given smaller batches of paragraphs rather than handling all the retrieved context at once. This phenomenon aligns with the well-known "lost in the middle" problem \cite{liu2024lost}, wherein models struggle to access relevant information situated in the middle of long contexts, even for models designed to handle long contexts. Consequently, this results in lower comprehensiveness scores, as some relevant information is overlooked despite using the same retrieved paragraphs in both setups.

\noindent \textbf{Automatic Evaluation}
We conduct an automatic assessment using the G-Eval framework across the four dimensions with the 10 sampled headings. The LLM is prompted with the generated content and headings, along with specific instructions tailored for each metric. To evaluate citation faithfulness, we include the original paragraphs from the cited case law judgments within the generated content. For comprehensiveness, we provide the actual content corresponding to each heading from the case law guide. As shown in Table \ref{tab:con_gen}, the automated assessments correlate closely with human evaluations across these dimensions. While expert assessment remains essential for gauging the quality and utility of structured reports, our findings indicate that automated LLM-based evaluations using the G-Eval framework can deliver rapid insights, offering a cost-effective alternative to expert assessments.

\subsection{Qualitative Case Study}
We conduct a qualitative case study on the LexGenie-generated report focusing on the `Rights to LGBTI Persons'. A complete generated report is provided in \url{https://tinyurl.com/43f86jw8}. The most compelling aspect identified is the detailed treatment of discrimination and equality rights, particularly the focus on intersectionality under sub-topic II.A2. This section effectively illustrates how LGBTI rights, though not explicitly enumerated in the European Convention on Human Rights, have been progressively built through interpretations of various articles, notably Article 8 (private life) and Article 14 (discrimination). These provisions have been instrumental in advancing LGBTI protections, including adoption rights, succession rights, marriage equality, and pension benefits. The system’s ability to highlight these key substantive aspects captures the ECHR’s approach to addressing discrimination against LGBTI individuals. Sections II and III provide the most insightful overviews, offering well-supported legal protections, and references to relevant case citations, supporting those claims.

Despite these strengths, the model has notable shortcomings. It overlooks crucial contextual insights, such as the role of states’ duties and positive obligations, which are vital for understanding discrimination cases. Additionally, it fails to address significant areas like migration issues, which span Articles 3, 8, and 5, and hate crime protections under Articles 3, 10, and 11. These omissions undermine a comprehensive understanding of the ECHR's jurisprudence. Structurally, the absence of transitional sub-topics or thematic connectors disrupts the logical flow, making it difficult to grasp the interconnected nature of topics like freedom of assembly (V) and LGBTI rights. This limitation stems from the current content generation pipeline, which focuses on isolated subsections without addressing cross-section redundancies or integrating detailed contextual links. Bridging these structural and contextual gaps could greatly enhance the usability and coherence of these generated guides.


\section{Conclusion}
In this paper, we introduce LexGenie, an automated LLM-based pipeline designed to generate structured report based on user-specified query from extensive case law, specifically within the ECHR jurisdiction. LexGenie’s two-stage pipeline first retrieves and organizes relevant passages according to user-defined topical queries, creating a structured outline that captures core legal issues and patterns. In the second stage, it generates cohesive, contextually accurate content for each section, providing a nuanced understanding of complex legal matters. Expert evaluations confirm LexGenie’s effectiveness in delivering relevant, well-organized reports, illustrating its potential to enable scalable, high-quality legal analysis. Additionally, initial automated evaluations using LLMs indicate a promising alternative to traditional expert reviews. Despite its strengths, challenges such as improving context integration and addressing structural flow remain.  Future work can expand to other jurisdictions and integrate multi-case analysis tools, such as temporal trend identification, to further support legal professionals in dynamic legal landscapes.

\section*{Limitations}
One key limitation lies in the quality of the retrieved passages and their clustering. Although the system aims to organize content into meaningful outlines, errors in retrieval or clustering can result in misaligned or overly broad sections that dilute the coherence of the report. This issue is particularly pronounced when dealing with ambiguous or overlapping topics, where the system may fail to distinguish fine-grained distinctions between related legal principles. Additionally, the pipeline does not currently incorporate mechanisms for ranking retrieved content by legal importance or authoritativeness, which can lead to the inclusion of peripheral or temporally outdated information \cite{santosh2024towards,santosh2024chronoslex}.

Another limitation is the lack of advanced contextual linking across sections. LexGenie generates content for individual subsections in isolation, which often results in a disjointed narrative that fails to capture the interconnected nature of legal issues. This fragmentation can hinder a comprehensive understanding of the broader legal landscape and reduce the utility of the generated reports for complex legal analyses.

\section*{Ethics}
We utilize case law data from HUDOC, the official database of the European Court of Human Rights. This publicly available data includes the real names of individuals involved, as the judgments are not anonymized. However, our work engages with this data solely for research purposes, without any intent or functionality that could exacerbate harm beyond the inherent exposure of the data's public availability.

LexGenie is developed as a tool to assist legal professionals by automating the generation of structured reports from case law, enhancing the efficiency of legal research. The system is intended to augment human expertise rather than replace it. While LexGenie provides valuable insights, its outputs may contain errors, such as hallucinated or misinterpreted legal references, which necessitate careful review and validation by qualified professionals. Users are explicitly advised against relying solely on the system for critical legal decisions. By ensuring the tool’s transparency and openly sharing its methodology, we aim to promote responsible use while underscoring the need for human oversight in all applications.

Additionally, the reliance on pre-trained large language models introduces the risk of perpetuating biases present in the training data. Legal judgments often reflect historical biases or systemic inequities, and there is a potential for these to be inadvertently amplified in LexGenie’s outputs. To address these challenges, we advocate for continuous monitoring, user feedback and iterative improvements to the system. This includes efforts to identify and mitigate any biases, ensuring that the tool aligns with ethical standards.

\bibliography{custom}

\appendix

\section{LexGenie Prompts}
\subsection{Keyphrase Generation}
Table \ref{tab:keywords_example} provides an example of paragraph and generated keyphrases. Prompt \ref{prompt:keyphrase} provides the prompt used for generating keyphrases. 
\label{keyphrase}
\subsubsection{Prompt}
\begin{figure*}[h]
  \centering
  \small
  \captionsetup{name=Prompt}
\begin{lstlisting}
You are an ECHR lawyer trying to create Legal Case-Law Guides that provide an in-depth overview of Convention ECHR case law on a particular Article or Transversal Theme. You will receive a paragraph extracted from case law; your task is to generate keywords that capture the essence of the paragraph so these keywords reflect the relevant Article or Transversal Theme and can be used to cluster cases, identify important cases, generate the table of contents and content for the Guides.

[Instructions]
1. Identify cross-references between paragraphs and reveal their connections;
2. Make sure keywords reflect the overall context of the paragraph by linking the description of circumstances to the requirements provided as criteria for legal doctrines and norms;
3. Map keywords like 'sometimes', 'exceptionally', 'in the present case' with the view to make sure that there is correspondence between legal standards and circumstances;
4. Focus on keywords detailing the application of substantive or procedural limb/branch explaining the scope of application of the Article;
5. Make sure to map accordingly keywords that detail the application of the Article to a variety of persons such as victims, state agents, witnesses, relatives, and similar;
6. Make sure to map accordingly keywords that detail the application of the Article depending on the jurisdiction, material, or temporal and those which detail the repartition or just satisfaction;
7. Distinguish conditions for the application of the Article in the context of violence/force from conditions that detail other events such as accidents or industrial activities;
8. Carefully identify key phrases that describe risks and operational choices from keywords that describe the creation and application of regulatory framework and conditions for responsibility of and accountability of various actors;
9. Highlight keywords that describe thresholds or conditions concerning intensity, frequency, and ordering in assessing each of the above.

[Paragraph]
{paragraph}

Please return ONLY the keywords for the given paragraph in one line and nothing else. Make sure to keep keywords in arguments together so they make sense.
\end{lstlisting}
  \caption{Generating keyphrases from paragraphs of case law judgements.}
  \label{prompt:keyphrase}
\end{figure*}

\begin{table}[htpb]
  \centering
  \begin{tabular}{p{2cm} | p{5.5cm}}
    \toprule
      Paragraph & The applicant submitted that the \colorbox{pink}{manner} in which he had been \colorbox{pink}{forced} to undergo the \colorbox{pink}{medical intervention} had amounted to \colorbox{pink}{torture}. The taking of the \colorbox{pink}{urine sample} had been \colorbox{pink}{coercive}, and he had never given his \colorbox{pink}{consent} to the \colorbox{pink}{procedure}. \\
    \midrule
      Keyphrases & \colorbox{pink}{forced medical intervention, coercive}, \colorbox{pink}{lack of consent, urine sample,} \colorbox{pink}{torture, manner of procedure.} \\
    \bottomrule
  \end{tabular}
  \caption[An example of a paragraph along with its generated keyphrases using Mistral-7B-Instruct model]{An example of a paragraph along with its generated keyword representations.}
  \label{tab:keywords_example}
\end{table}

\subsection{Structure Generation} \label{strcuture}
Prompt \ref{prompt:topic} and \ref{prompt:organize} provide detailed prompts used for generating topic name for each cluster and final LLM call to organize the generated topics and sub-topics into a coherent table of contents respectively.
\begin{figure*}[h]
  \centering
  \small
  \captionsetup{name=Prompt}
\begin{lstlisting}
You are given a list of paragraphs extracted from the European Court of Human Rights case law, and your task is to generate a detailed topic label to represent these paragraphs in ECHR case law guidelines. Here is the list of paragraphs:

[DOCUMENTS]

Based on the information above, generate a detailed topic label in the following format and nothing more:
topic: <topic label>
    \end{lstlisting}
  \caption{Generating topic name for each cluster.}
  \label{prompt:topic}
\end{figure*}

\begin{figure*}[h]
  \centering
  \small
  \captionsetup{name=Prompt}
\begin{lstlisting}
I have a list of topics related to European Court of Human Rights (ECHR) case law documents. I would like you to organize these topics into a coherent and structured Table of Contents (ToC) similar to a legal document ECHR guidelines. Please group related topics under appropriate sections and subsections, ensuring a logical flow. The ToC should include main headings, subheadings, and possibly further subdivisions where necessary with 4 spaces indentation and without general sections such as introduction and conclusion. The final structure should resemble an outline for comprehensive legal report guidelines that align with the topics from ECHR. Here is the list of topics:

[Topics]
{Topics}

Please only return a well-structured ToC and nothing else.
\end{lstlisting}
  \caption{Organize topics into a hierarchical structure.}
  \label{prompt:organize}
\end{figure*}

\subsection{Content Generation}
\label{content_gen}
Prompt \ref{prompt:content} provides the prompt for iterative content generation approach for each leaf sub-section in the table of contents.
\begin{figure*}[h]
  \centering
  \small
  \captionsetup{name=Prompt}
\begin{lstlisting}
You are a legal expert tasked with generating content for a Case Law Guidelines section based on the given section heading, current section content, and a set of paragraphs extracted from case law documents. Your goal is to synthesize the information from these paragraphs to extend and create clear and accurate content without sections like introductions or subsections. The content should be strictly related to the heading and logically coherent, and the relevant paragraphs from the case law documents should be cited by their IDs. Provide thorough explanations, elaborate on key points, and include examples where relevant. Follow the instructions below carefully to ensure the guidelines are precise and informative.

[Instructions]
1. Review the provided set of paragraphs extracted from case law documents;
2. Consider only those paragraphs that are strictly related to the keywords in the heading;
3. Develop content based on the information principles contained in the paragraphs and ensure the content is clear and concise;
5. Citations: whenever a guideline is influenced by or derived from a specific paragraph, cite that paragraph by its id and number in parentheses as (id#paragraph_number);
6. Maintain a professional and formal tone throughout;
7. Only generate the content in relation to the keywords in the heading and focus on the specific standards implied by those keywords;
8. Return a coherent answer comprising general observations and standards from the Convention and specific observations and standards implied by the keywords in the heading;
9. Extend the previously generated content with the new content, revising and integrating it smoothly to form a coherent narrative;

[Heading]
{Heading}

[Previous Content]
{Previous_Content}

[Paragraphs]
{Paragraphs}

Return the generated content and nothing else. Make sure to use only the related paragraphs to the heading.
[Your response]
\end{lstlisting}
  \caption{Content Generation.}
  \label{prompt:content}
\end{figure*}

\section{LexGenie UI}
\begin{figure*}[h!]
    \centering
    \begin{subfigure}[t]{\textwidth}
        \centering
        \fbox{\includegraphics[width=\linewidth]{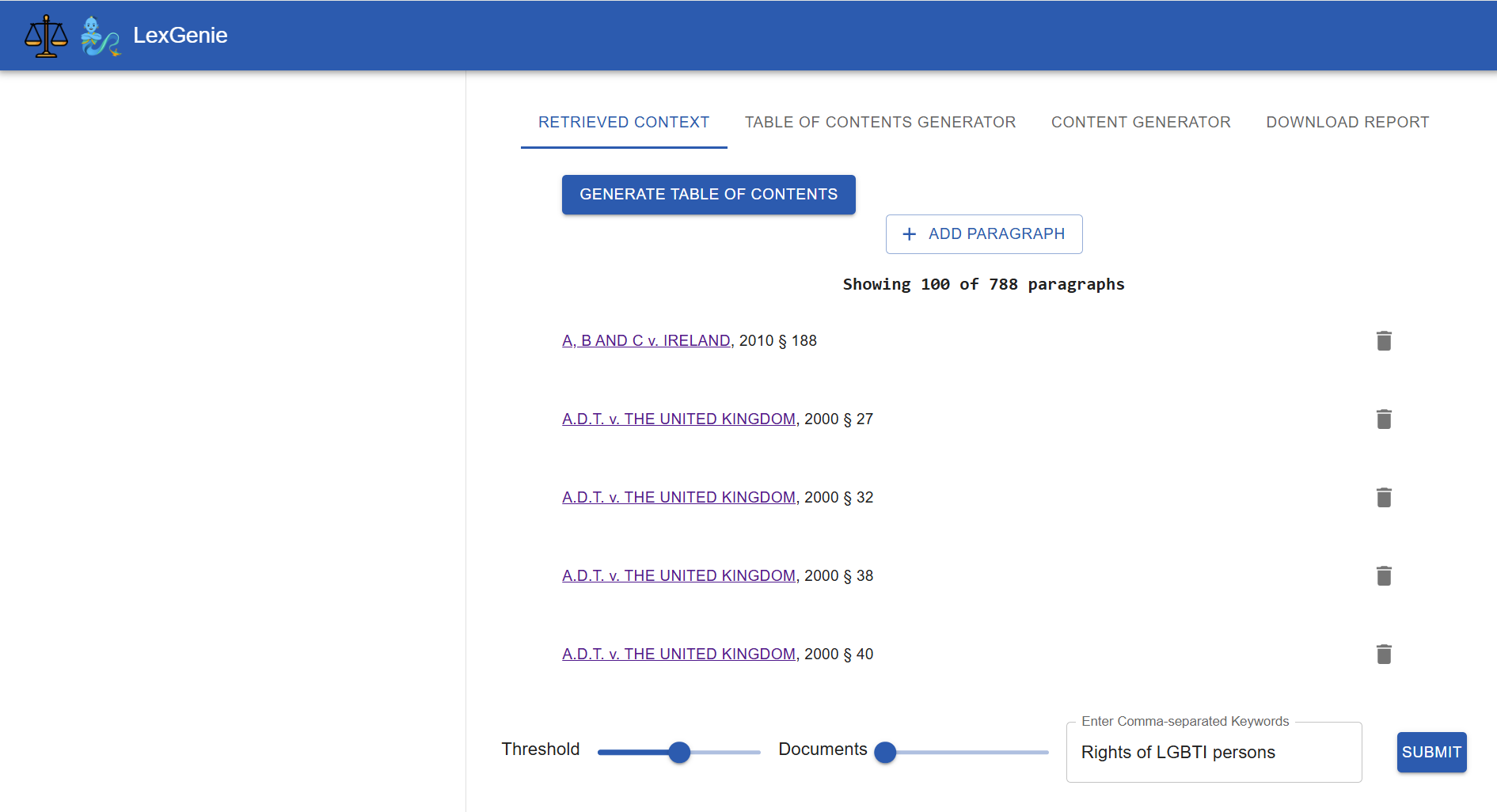}}
        \label{fig:image1}
    \end{subfigure}
    \hfill
    \begin{subfigure}[t]{\textwidth}
        \centering
        \fbox{\includegraphics[width=\linewidth]{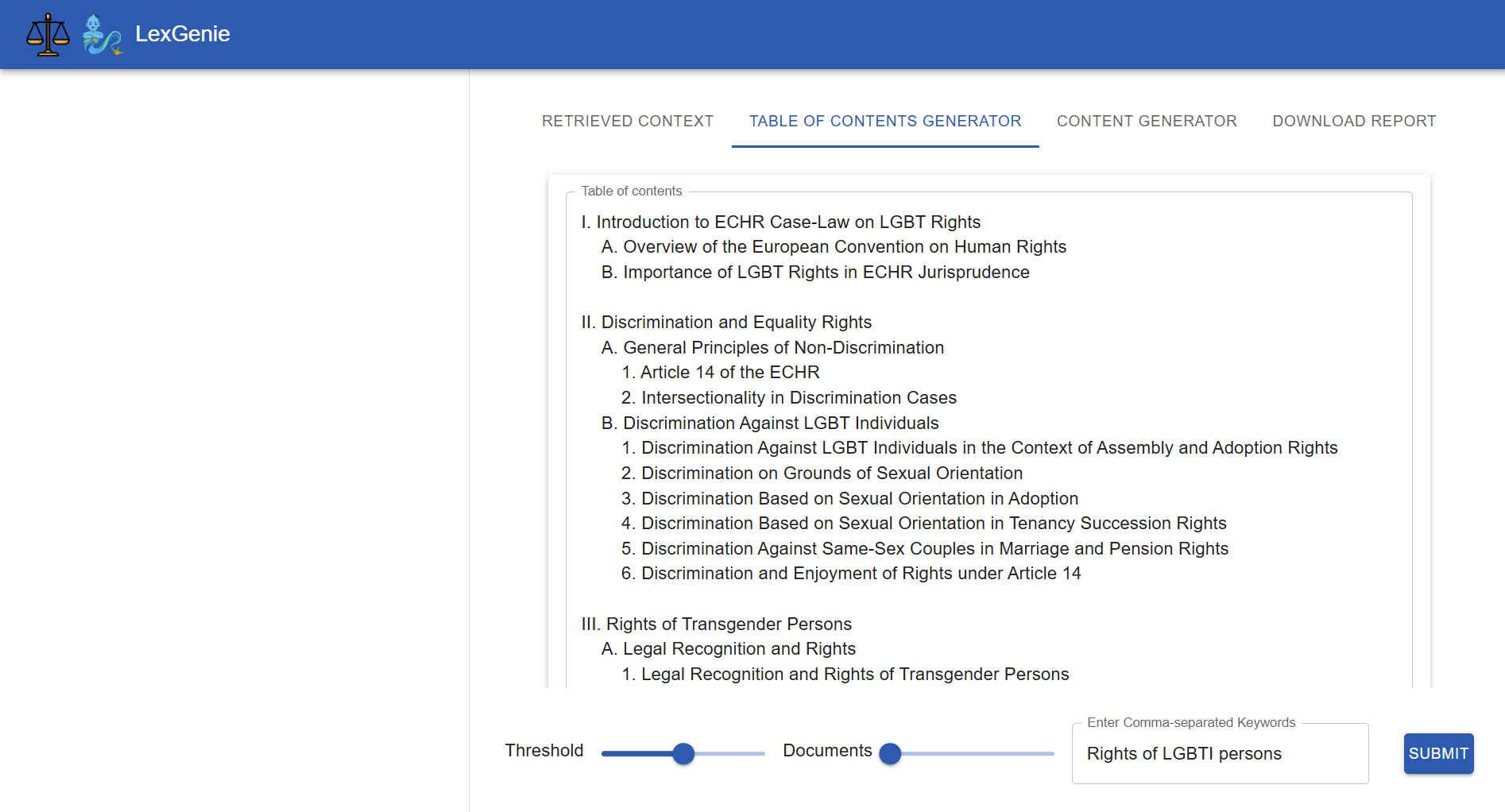}}
        \label{fig:image3}
    \end{subfigure}
    \caption{LexGenie interface. Given a legal topic as query, it automatically retrieves relevant documents and generates a table of content structure for the report. Finally, content for each sub-section in report is populated and the whole report is available for download.}
    \label{app:ui}
\end{figure*}

Figure \ref{app:ui} displays the UI interface and functionalities offered through LexGenie.

\end{document}